# AI-Enhanced Intensive Care Unit: Revolutionizing Patient Care with Pervasive Sensing

Subhash Nerella[1,2], Ziyuan Guan[2,4], Scott Siegel[1,2], Jiaqing Zhang[2,3], Ruilin Zhu[2,4], Kia Khezeli[1,2], Azra Bihorac[2,4], Parisa Rashidi[1,2]

[1]Department of Biomedical Engineering, University of Florida, Gainesville, Florida, USA
[2]Intelligent Critical Care Center, University of Florida, Gainesville, Florida, USA
[3]Department of Electrical and Computer Engineering, University of Florida, Gainesville, Florida, USA
[4]Department of Medicine, University of Florida, Gainesville, Florida, USA

Abstract

The intensive care unit (ICU) is a specialized hospital space where critically ill patients receive intensive care and monitoring. Comprehensive monitoring is imperative in assessing patients' conditions, in particular acuity, and ultimately the quality of care. However, the extent of patient monitoring in the ICU is limited due to time constraints and the workload on healthcare providers. Currently, visual assessments for acuity, including fine details such as facial expressions, posture, and mobility, are sporadically captured, or not captured at all. These manual observations are subjective to the individual, prone to documentation errors, and overburden care providers with the additional workload. Artificial Intelligence (AI) enabled systems has the potential to augment the patient visual monitoring and assessment due to their exceptional learning capabilities. Such systems require robust annotated data to train. To this end, we have developed pervasive sensing and data processing system which collects data from multiple modalities depth images, color RGB images, accelerometry, electromyography, sound pressure, and light levels in ICU for developing intelligent monitoring systems for continuous and granular acuity, delirium risk, pain, and mobility assessment. This paper presents the Intelligent Intensive Care Unit (I2CU) system architecture we developed for real-time patient monitoring and visual assessments.

## 1. Introduction

Intensive Care Units are specialized treatment facilities in a hospital where critical care and life support are provided to acutely ill patients. Recognizing declining clinical status through frequent patient assessments [1, 2] is a cornerstone of ICU care, saving up to $1 billion in terms of quality-life-year gained [3]. Although close monitoring and dynamic patient acuity evaluation are key components of ICU care, both are constrained by the time restraints placed on healthcare professionals. Existing clinical acuity indices are developed for mortality prediction, use low-frequency data, lack automation, human observation-based, and have limited accuracy and interpretability [4, 5]. Consequently, clinicians must rely on their clinical judgment and vigilance for patient acuity assessment. Moreover, the information related to pain, emotional distress, and physical function that requires visual observation of a patient by the nurse or patient self-report are captured

sporadically [6, 7]. Real-time close monitoring can be advantageous for acuity evaluation, delirium risk prediction and assessing pain.

Delirium is a frequently encountered issue among patients receiving care in the ICU. Delirium is characterized as a serious deviation from normal mental abilities due to acute brain dysfunction, with incidence rate ranging from 45% to 87% [8] in the ICUs. Delirium is associated with prolonged hospitalization, an increase in mortality, and long-term cognitive impairment risk [9-11]. The risk factors associated with delirium are not routinely monitored, such as patient immobility, improper noise, and light levels in the ICU.

Pain is a prevalent phenomenon observed in individuals admitted to the ICU. Despite efforts by clinicians and ICU personnel, evaluating pain in non-verbal, sedated, mechanically ventilated, and intubated patients remain a challenging task. Patient self-reported pain scores and nurse observations suffer from subjectivity, poor recall, limited frequency, and lag between observation and documentation, potentially leading to delayed interventions [12-14]. ICU staff currently rely on routine manual visual examination to assess functional status, pain, and other critical care indices [15]. These repetitive measurements increase the burden on ICU personnel and lead to burnout [16], with almost a third of ICU nursing teams reporting burnout [17]. A high nursing workload contributes to a higher incidence of adverse events in the ICU [18-21].

ICU nurses and physicians spend significant time interacting with Electronic Health Records (EHR), either documenting or retrieving data impacting the overall time they spend on direct patient interactions [22-24]. Repetitive manual observations by overburdened ICU nurses are prone to documentation errors [25]. Hence, there exists an urgent need to expedite and automate routine monitoring tasks in the ICU [26]. Furthermore, present protocols for visual assessments of patients do not capture subtle changes in facial expression, patient body posture, and mobility. These important visual cues, if collected, can be associated with pain experienced, physical function, and emotional distress. There is a critical unmet need for real-time, dynamic, interpretable, and precise visual assessment of the patient for acuity and delirium risk prediction.

In this paper, we present the I2CU system architecture we developed for real-time pain and mobility assessments, monitoring sound pressure levels, light intensity levels, and visitation frequencies, which enable the development of AI-enhanced decision support tools including patient acuity monitoring, pain assessment and delirium risk quantification. In the following sections, we discuss the data collection modalities and sensor devices uses, working of our real-time I2CU system and major components including software and hardware, data curation steps, automated data processing pipelines for generating data for annotation, face, and depth annotation along with user interfaces, and active learning pipelines.

## 2. Study Participants

All the data utilized in this study was obtained from adult patients admitted to the surgical ICUs at the University of Florida Shands Hospital in Gainesville, Florida. We complied with all applicable federal, state, and local laws and regulations. The study was approved under IRB202101013, IRB201900354, IRB201400546, and IRB202101147 by the University of Florida Institutional Review Board. We obtained written informed consent from all patients before enrolling them in the study. For patients who were unable to provide informed consent, we obtained consent from a legally authorized representative on behalf of the patient. Patients over 18 years of age admitted to ICU and expected to remain in the ICU for at least 24 hours were eligible for recruitment. Exclusion criteria include patient transfer, discharge, and death within 24 hours of recruitment. Also, patients who require any form of isolation, contact precaution, or the absence of informed consent (either by the patient or legal representative) were excluded from the study.

### 2.1 Data Collection

We collected various modalities of data from ICU including clinical, imaging, wearable, environmental, and physiological data shown in Table 1. Clinical data includes patient medications, lab results, demographics, and critical care indices. Imaging data includes RGB videos of patient faces and depth images with a view of the entire ICU room. Physiological data includes vital signs, heart rate, blood pressure, temperature, and respiratory rate. Wearable data includes accelerometer data, gyroscope, and electromyography (EMG) data. Environmental data comprises noise, light, and air quality in the ICU. We collect patient's data for a duration of seven days, or until the patient completes the study, whichever is shorter.

| Modality | Data type | Sensor | Sampling Frequency | Target |
|---|---|---|---|---|
| RGB | Image | Amcrest | 1 FPS | Patient face |
| Depth | Image | Azure Kinect | 1 FPS | ICU room |
| | | Intel Realsense L515 Lidar | 1 FPS | |
| Accelerometry | Time series | Actigraph GT3X | 30 hz | Patient limb extremities |
| | | Shimmer 3 EMG unit | 100 hz | |
| EMG | Time series | Shimmer 3 EMG unit | 512 hz | Bicep |
| Noise | Time series | Thunderboard EFR32BG22 | 1 hz | Noise levels in ICU |
| Light | Time series | Thunderboard EFR32BG22 | 1 hz | Light levels in ICU |

Table 1. Sensor devices used in the Intelligent ICU system, the sampling frequency of data collection, and the target location for each sensor.

## 3. Intelligent ICU System Architecture

The Intelligent ICU system architecture depicted in Fig. 1 enables the reliable and continuous collection of data from sensors and EHR, perform data processing, run Machine Learning (ML) models, and provide an interpretable visualization of model outputs and predictions for care providers. The major components of the I2CU system are (1) Sensors, (2) The pervasive cart system with Nvidia AGX Orin toolkit for processing the data and running ML algorithms (3) The central server that hosts the web server and web application to control the sensors, data storage, and GPUs to train ML models.

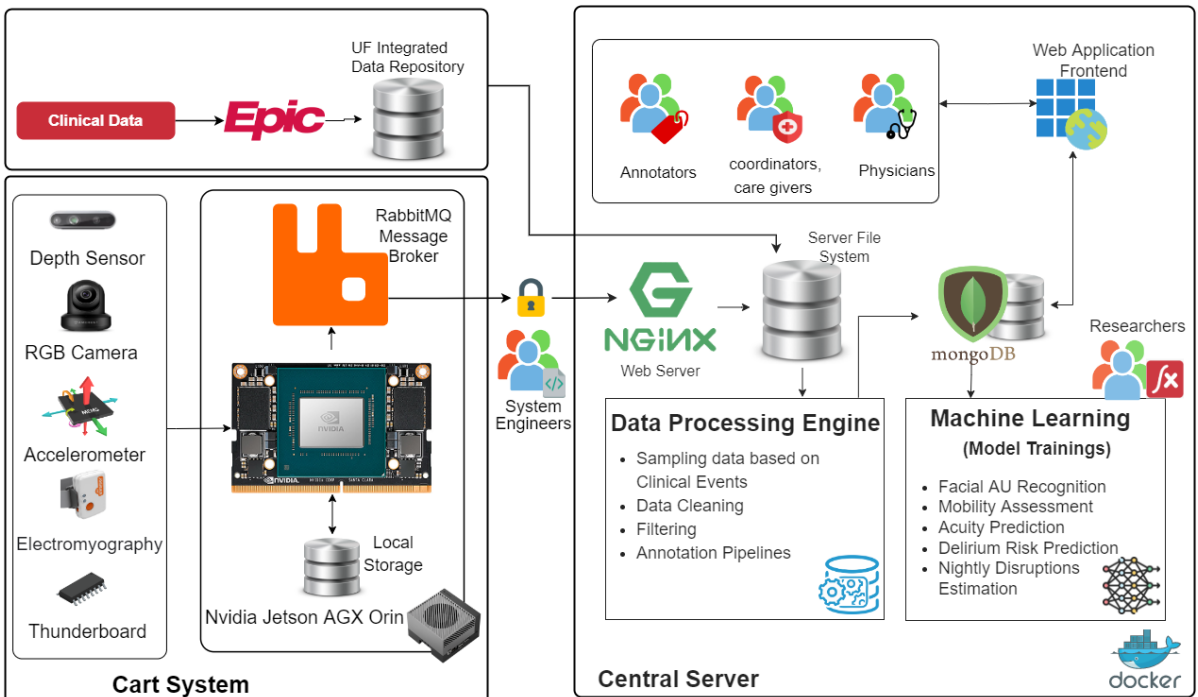

Fig. 1. Schematic of Intelligent ICU system architecture. Imaging sensors mounted on the cart system record data and stored it in the local storage. The data is uploaded through a secured VPN connection to a secured server on the UF health network.

### 3.1 Pervasive Sensing System

We have built six portable pervasive sensing cart systems that can be wheeled in and out of the ICU. The cart is mounted with an RGB camera with pan, tilt, and zoom functionality, a depth camera looking at the entire ICU room, and a Thunderboard with light, noise, and environmental sensors. We have also developed a user interface for the ICU nurses and clinical coordinators, to start, stop, and pause the recording during medical procedures or upon a patient's request. The recording control can be directly from the cart and remotely.

## 3.2 Server Machine

We use a dedicated server machine to store, curate, and train machine learning models on data collected in the ICU. We chose to have a dedicated central server machine for several reasons:

1. Patient data are complex and obtained in multiple modalities, requiring large data storage and processing capabilities. On average, the data collected from each patient during their ICU stay require 400 Gigabytes of storage.
2. Patient data is confidential and must be protected from potential data breaches. The server is on the UF health network and can only be accessed through secured UF health VPN connection ensuring data security and privacy.
3. Training machine learning models on patient data demands significant computational resources. Server machines can host multiple GPUs speed up the training process and make training models more efficient. Our server is equipped with three Nvidia 2080Ti GPUs.
4. Trained machine learning models need to be validated and tested before deploying into production for real-time applications. Machine learning models on multiple stages of deployment are hosted on the server machine.
5. The server can host the webserver, and the web application to control the data collection on multiple carts remotely enhancing the scalability of the system.

## 3.3 Real-time Monitoring

Prompt observation of the change in patient clinical status in the ICU is a key input for decision-making for physicians [27]. To capture the fast-changing clinical status of the patient the system should be capable of handling the request and data flow in real-time without any losses. The schematic diagram of the end-to-end data and request flow is shown in Fig. 2 between the pervasive cart system and the central server. The major hardware components of the cart system are sensors and the Nvidia Jetson AGX Orin dev kit. The central server collects the data streams from multiple carts, hosts a web server and a web application to control the sensors on the carts, and hosts a web-based dashboard application to visualize the patient's clinical status.

### 3.3.1 Data Processing Module

The data processing module includes metadata tagging, data compression, and encryption. We include a timestamp of the data collection, cart identifier, ICU room identifier, and individual sensor identifier to the data collected from the sensors as metadata for data curation that happens at a later stage on the central server. The data collected is compressed to reduce data transmission bandwidth to be sent to the central server. The data is also encrypted to ensure the security of patients' Protected Health Information (PHI).

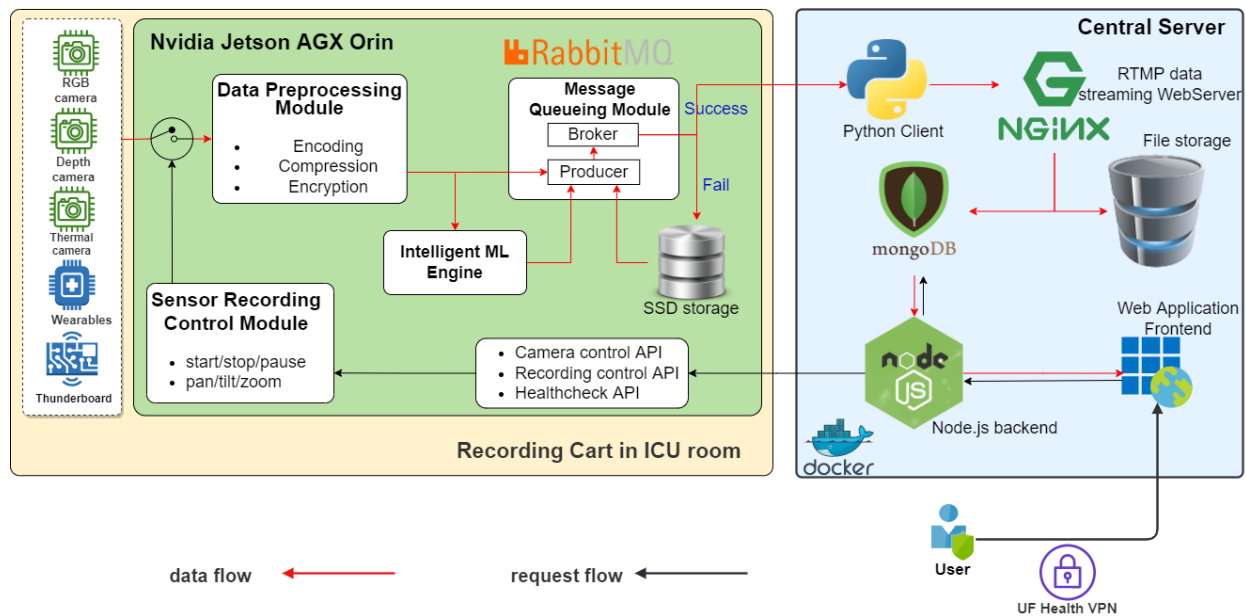

Fig. 2. A schematic figure showing data and requests between the pervasive cart system and the central server. A web application deployed on the central server can be used to control the sensors on the cart system. The Nvidia Jetson consists of multiple modules for processing, ML inference, and queueing data.

### 3.3.2 Intelligent ML Engine

The intelligent ML engine module comprises multiple end-to-end deep learning pipelines deployed on the Nvidia AGX Orin device, running real-time inference on sensor data and EHR data. The tasks performed in the ML engine include:

1. Visual assessment of patient faces, we extract frames from a real-time video feed of RGB data at 1 FPS. We perform face detection and crop the faces using MTCNN [28] model. These cropped faces are fed to a trained Swin-transformer [29] for detecting facial action units on the faces.
2. Visual assessment of mobility, we extract depth image frames at 1 FPS from the depth camera. YOLO V5 model trained on the ICU-mobility dataset is inferred posture on the depth frames.
3. Brain acuity prediction using a trained Self-supervised Transformer for Time-Series (STraTS) [30] model on over 300 clinical variables obtained from IDR at regular intervals,
4. Sound pressure and light levels in the ICU
5. Visitation frequency during the day and night.

### 3.3.3 Messaging Queue

The I2CU system comprises of multiple data-collecting components, including sensors and multiple data sources from the IDR. A data producer gathers all the streams of data from different sources for the downstream tasks. The data producer is critically important in the case of our data-driven system for the timely availability of data on the central server to ensure system health checks of the sensors, control the cameras, and visualize the risk score generated based on the ML model inferences. We use RabbitMQ, an open-source message queuing tool for data handling purposes. We run an instance of RabbitMQ on the Nvidia Orin AGX dev kit as a producer and broker. With RabbitMQ, we could ensure zero data loss even during unstable network connection or network failure by caching the data into local storage and retry the data transfer at a later time. We use separate exchanges and queues to route the data obtained from different sensors. RabbitMQ communicates with the central server using a transport layer security (TLS) certificate.

A python client running on the central servers collects the data from multiple cart systems publishing data through the RabbitMQ messaging queue service. The data is fed to the nginx web server that supports real-time messaging protocol (RTMP) to stream the data collected from the server and simultaneously store the data on the central server. The RTMP data stream is used to visualize ML model inference and for controlling the sensors on the cart, which include start/stop/pause the data collection and camera orientation.

### 3.3.4 I2CU Web Application

Our I2CU system is developed with capabilities to control the sensors through a secure web application. This web application is run on a Docker container connected to the RTMP data stream from the nginx webserver. The application is developed using Node.js express platform to fetch data from the RTMP data stream, and React library is used to build the user interface shown in Fig. 3. The user interface shows a preview of all the carts deployed in different ICU rooms and the health status of the cart, and the sensors mounted. Clinical coordinators will be able to control the camera recording on the cart and adjust the orientation of the RGB camera remotely as depicted in Fig. 4. As the patient's face needs to be captured by the RGB camera, it is essential to periodically check the orientation of the camera to ensure data quality.

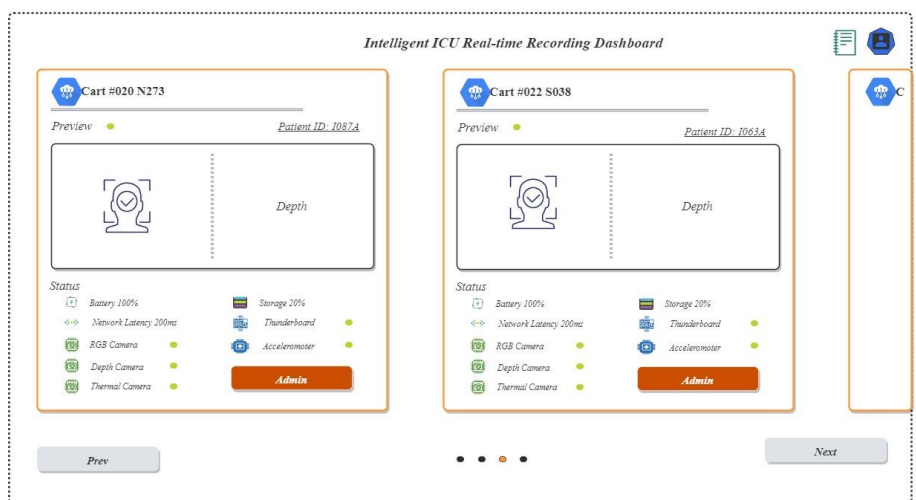

Fig. 3. Intelligent ICU web application user interface showing data stream preview from the carts individually

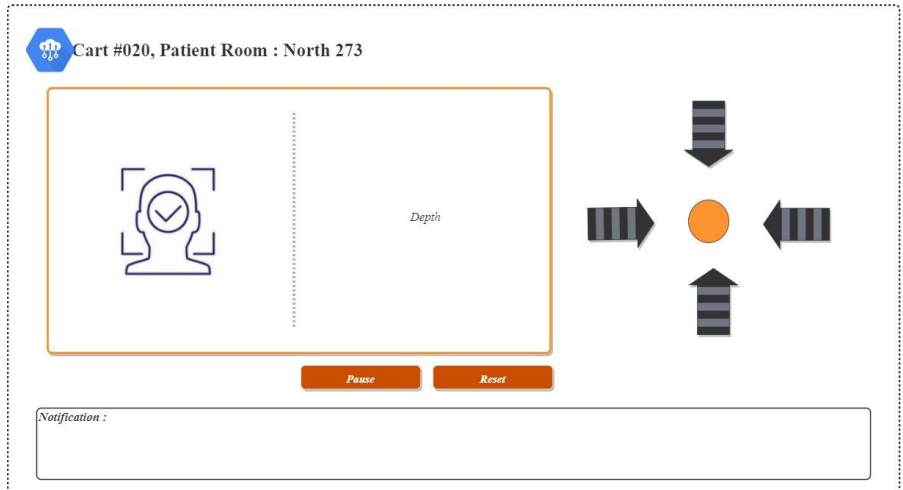

Fig. 4. Cart control user interface to control the camera orientation and start/stop/pause the data collection.

### 3.3.5 Dashboard

The dashboard provides clinicians with a quantitative score for patient clinical status and the severity of the illness, further assisting in clinical decision-making. The dashboard shows patient acuity, delirium risk scores, pain levels, and mobility, in addition to ICU room conditions including noise levels, light levels, and nightly disruption. High noise levels and bright lights in ICU are linked to delirium and sleep deprivation, while low mobility can increase risk and cause further complications. The dashboard user interface shown in Fig. 5 facilitates physicians to monitor these factors and make adjustments to the patient's environment and treatment plan to optimize the recovery plan and prevent adverse outcomes. The dashboard user interface was developed using React, a JavaScript library and Node.js express platform that fetches machine inference data stream from the nginx RTMP webserver.

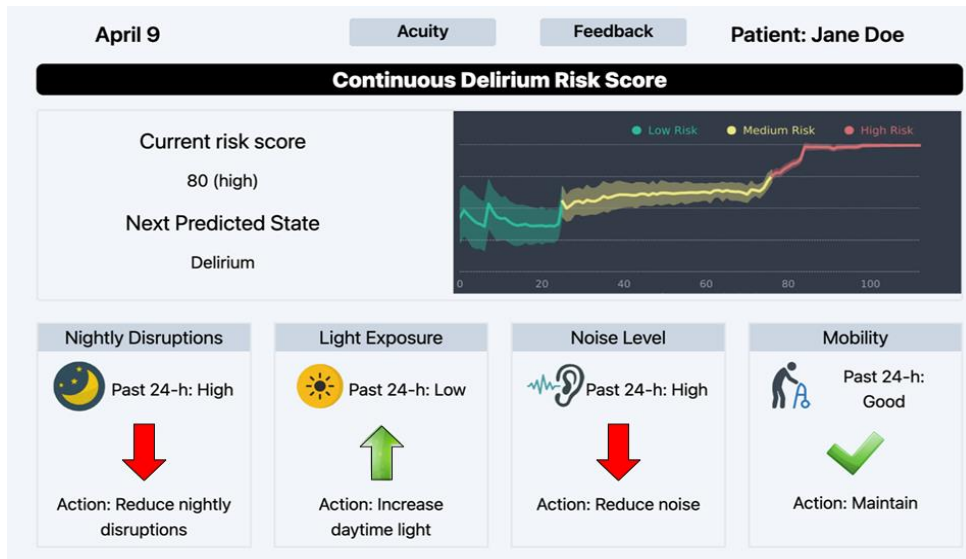

Fig. 5. The user interface of the I2CU dashboard shows patient acuity, delirium risk, nightly disruptions, light exposure, noise level, and mobility for the physicians.

## 3.4 Data Curation

The data curation process involves removing protected health information of the patient and separating the data specific to a patient based on admission and discharge date and time. Patient data is organized into modality-specific folders ensuring all patients' data is organized in the patient-specific folders on the server.

## 3.5 Data Processing Pipelines

We collected face and depth videos for seven days for all the patients recruited to the study. The data collected for each patient contains approximately 200 Gigabytes of face videos and 150 Gigabytes of depth images. A manual review of these videos to generate data for annotation is time-consuming. Therefore, we have developed automated data processing pipelines to process the recorded data.

The face image extraction pipeline shown in Fig. 6 starts by filtering videos within one hour of the patient-reported pain score timestamp. Pain time stamps are obtained from the UF IDR. We use FFmpeg [33], a multimedia processing tool, to extract image frames from the filtered videos, followed by multitask cascaded convolutional network (MTCNN) [28] to detect crop faces from the frames. We discard the frames whenever MTCNN outputs zero detection or detects more than one face in the frame. Image frames with more than one face are discarded because they contain faces of individuals, likely family or caregivers.

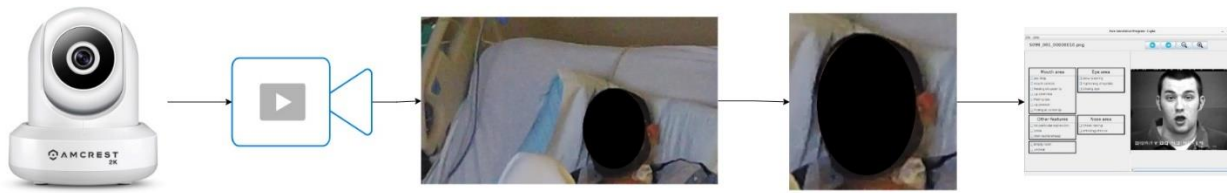

Fig. 6. Data processing pipeline for face annotation. The recorded videos are processed to extract frames. Patient faces are cropped from the frame, then cropped faces are annotated by trained annotators.

Depth image processing pipeline Fig. 7 starts with depth images stored as compressed arrays on the server. We have trained a YOLO V5 [31] to detect people in the depth frame, and only images that contain at least one person detected were retained for annotation while the rest were removed from the annotation queue. We use the scikit-image [34] structural similarity index to compute the similarity between successive images to minimize the redundant annotation.

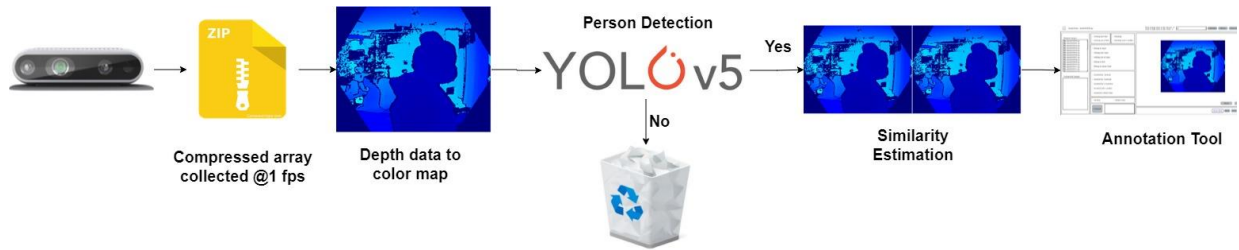

Fig. 7. Depth image processing pipeline includes converting compressed depth arrays to colormaps. The YOLO V5 model is used to filter images containing people and similarity estimation between sequence images to minimize redundant annotation.

### 3.6 Data Annotation

Annotated data is the core of supervised deep learning as it provides the information necessary for training models in order to make inferences on unseen data. In general, deep learning models need a large and diverse annotated dataset for robust learning and confident predictions. As a part of our I2CU system, we annotated face and depth images. We have developed secure, user-friendly annotation tools with a user interface to annotate the imaging data. To ensure the security of patient data and customize the annotation specific to our study, we have developed in-house annotation tools. Annotators were asked to register in the annotation tool with their preferred usernames and password. The annotation tool frontend was built using the React framework. The frontend relies on a FastAPI backend server to communicate with a MongoDB database. The tools are capable of handling multiple annotators simultaneously performing annotation. Submitting an image annotation triggers a fetch request that FastAPI uses to update the MongoDB database. We have equipped the annotation tools with functionalities that can reduce the annotation error and ensure the reliability of the data for model training. These functionalities include the ability to come back to skipped images, add comments, and keep track of the number of images annotated in a session and a week.

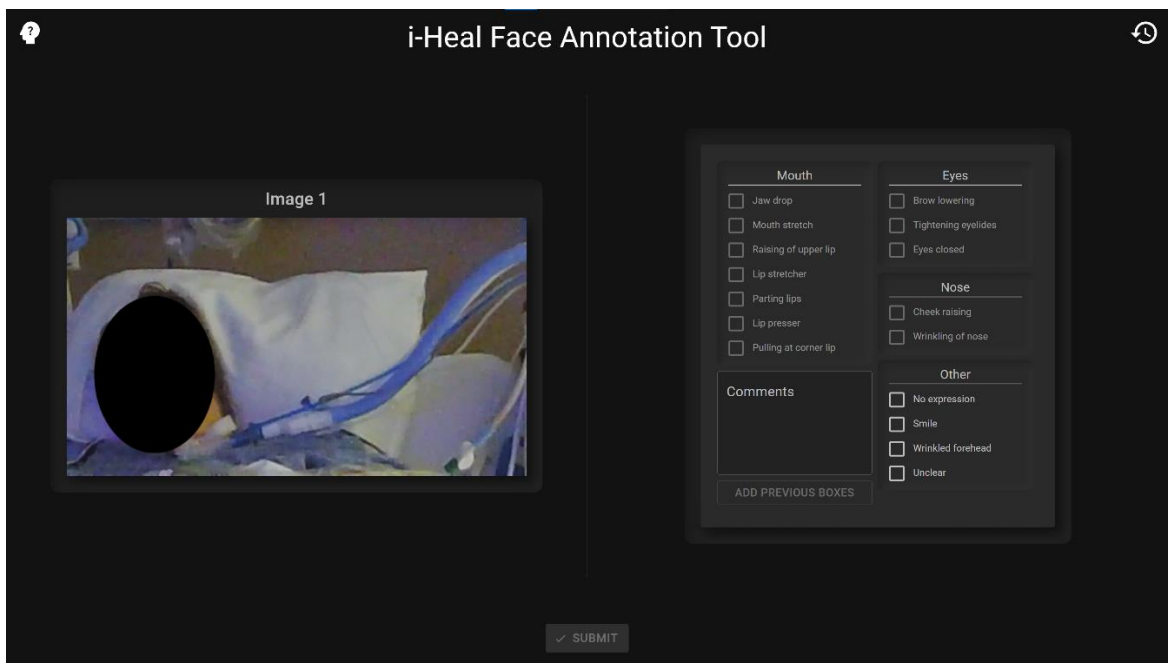

Fig. 8. Face annotation tool user interface. The annotation includes FACS system facial AUs, smile, unclear, and wrinkled forehead.

### 3.6.1 Face annotation tool

Many researchers approached facial action unit detection problems over the last decade. Although many facial action unit (AU) datasets got available over the years, all these datasets are collected in controlled or semi-controlled settings. In our previous work [30], we showed models trained on data collected in controlled environments are inadequate for AU detection in ICU patients. Unlike the controlled settings, ICU is a highly uncontrolled dynamic environment due to the presence of life support devices on patient faces, varying light intensity levels, altering patient face orientation, and partial face occlusions.

Our patient facial image annotation comprises 12 facial AUs defined by the Facial Action Coding System (FACS) [35] shown in Table 2, along with other facial features such as smile, wrinkled forehead, and no particular expression. The AUs annotated are chosen to include AUs from Prkachin and Solomon Pain Intensity (PSPI) score [36], and non-verbal pain assessment tools which include Non-Verbal Pain Scale (NVPS) [37], Behavioral Pain Scale (BPS) [38], and Critical Care Pain Observation Tool (CPOT) [39]. Patient faces are shown on the left half of the user interface shown in Fig. 8.

| AU | Description |
|---|---|
| 4 | Brow Lowerer |
| 6 | Cheek Raiser |
| 7 | Lid Tightener |
| 9 | Nose wrinkler |
| 10 | Upper Lip Raiser |
| 12 | Lip Corner Puller |
| 20 | Lip Stretcher |
| 24 | Lip Pressor |
| 25 | Lips part |
| 26 | Jaw Drop |
| 27 | Mouth Stretch |
| 43 | Eyes Closed |

Table 2. List of facial action units we annotate and their corresponding description

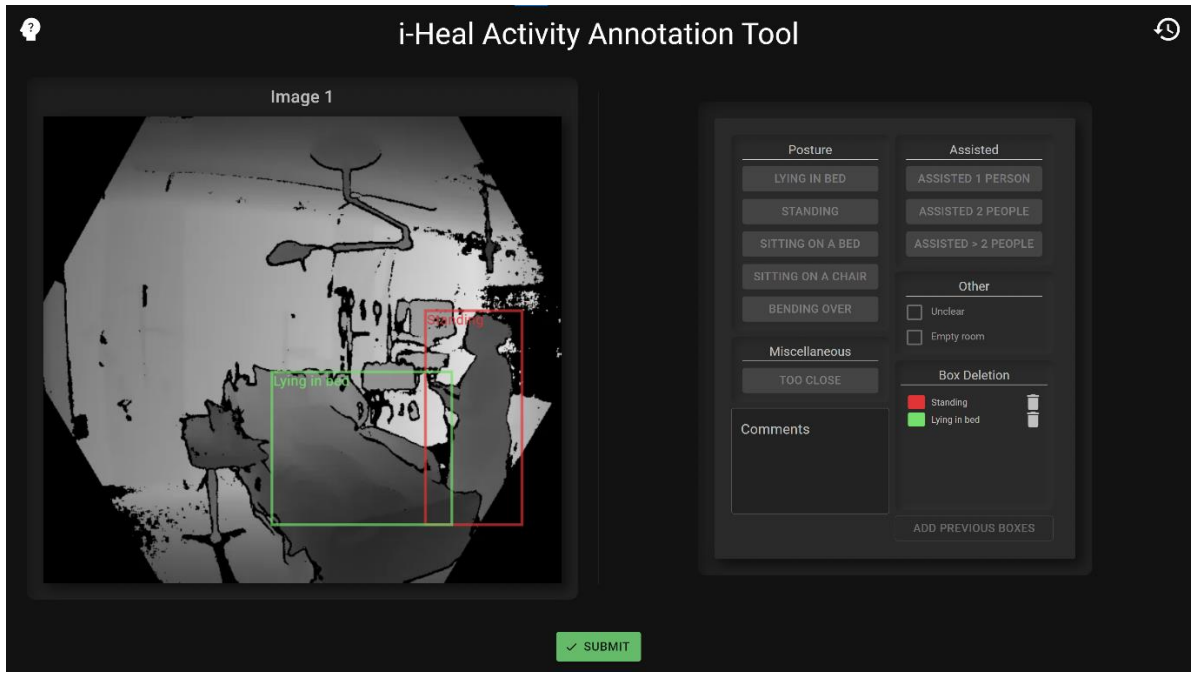

Fig. 9. Depth annotation tool user interface. Depth annotation includes posture, mobility, and assistance annotation. Depth annotation tool has draw box functionality to draw bounding box around the target region of the image.

### 3.6.2 Depth Annotation Tool

Patient mobility plays an important role in patient recovery in the ICU. Patient mobilization is shown to decrease the length of ICU stay, reduce the duration of mechanical ventilation and weaning, delirium onset, prevent muscle atrophy and improve functionality [40-42]. To utilize the data collected from depth cameras to develop ML models for activity recognition, we developed a custom secure annotation tool with a user interface. Three annotators were recruited and trained on sample depth image data from the ICU to perform annotation on the depth image frames. We divided the annotation into three categories patient posture label and assistance level. Posture labels include sitting and standing; assistance labeling comprises assisted by one person, assisted by two persons, assisted in a wheelchair, and assisted with a walker. The annotation tool is linked to the MongoDB database to record the data annotation and provide weekly summarization reports. We treat mobility, posture, and assistance detection as object detection problems. For the same reason, our annotation tool shown in Fig. 9 was designed to draw multiple bounding boxes on the image. Every box drawn will have a unique label that belongs to one of the three annotation categories.

### 3.7 Annotation Challenges.

We use AUs defined by the FACS system to annotate facial behavior. FACS annotation by itself is challenging as human expressions are incredibly complex, involving multiple facial muscle movements simultaneously, and breaking down the expression into individual action units accurately is difficult. Facial expression is subjective. In many cases, a given facial expression can be perceived differently by different individuals. Cultural differences can impact facial expressions, and particular expressions may not mean the same thing across multiple cultures. Anatomical differences also introduce variations in the appearance of facial expressions. Beyond these reasons, FACS annotation on patients in ICU poses additional challenges, which include the presence of life support devices such as tubes, ventilators, wires, and medical equipment that can obstruct patient faces. Patients in ICU are most likely to be under sedatives, painkillers, and other medications that can affect facial expressions. Varying light levels and inconsistent facial orientations pose additional challenges.

We use depth cameras to capture patient mobility and the assistance level needed. Although depth cameras are privacy-preserving, we also lose the ability to differentiate a patient from staff and visitors in many cases. Depth cameras cannot always resolve a patient lying in bed, noisy pixel outlier values impact colormap visualization during annotation and depth image normalization during model training.

### 3.8     Annotation Summary

Regular summaries of annotation progress are essential for effective data labeling to train machine learning models. Annotation summaries enable us to check on the labeling progress, timeliness of annotation, and efficiency. We compute annotation summaries on a weekly basis to obtain insights into the total number of annotations completed, hours of time taken by individual annotators, and the median time taken to annotate a single image. We also compute annotation agreement between annotators using the Fleiss Kappa score [43] for the week and also the entire dataset annotated. Weekly annotation summaries enabled us to make informed decisions regarding project priorities, budget allocations, hiring human resources, and quality control.

### 3.9     Active Learning Framework

Labeling data for facial AUs and mobility is expensive, time-consuming, and involves a steep learning curve. Our datasets include over 18 million depth image frames and 22 million patient face image frames extracted from videos. It is not practical to annotate the entirety of these massive datasets. Active learning is an important machine learning technique that involves an iterative process to choose most informative data samples to be labeled. The goal of active learning is to maximize ML model performance with minimum number of labels data points. As new patients are recruited, the data distributions change over time. Therefore, it is crucial to train the data on diverse patients to allow the model to adapt to changing distributions. The active learning framework is designed to identify relevant data samples to be annotated. Another important aspect is the annotator quality, which can significantly impact the training effectiveness of the machine learning model. Annotators should be evaluated for accuracy and consistency to ensure that the annotation is performed accurately and consistently across the dataset. The annotator label quality is computed based on the prediction confidence of the model for a given annotator's label and the annotator's agreement with the other annotators. We also compute consensus label quality score per sample and class, which is further used to calculate active learning scores for annotated samples. The active learning score is computed as a combination of the consensus label quality scores and ensemble of multiple machine learning models in the case of annotated samples to determine if the sample needs further annotation to increase the consensus label quality score. In the case of unlabeled samples, the active learning score is computed solely based on the ensemble models predicted confidence. We use the expected calibration error [44] to compute each model's weight on the active learning score of a given sample. Our approach is based on the Cleanlab [45] implementation of active learning for data annotation. We have adapted it to our facial annotation, a multilabel multi-annotator problem, by computing active learning scores per AU. For depth annotation, a multi-annotator object detection problem active learning scores are computed per box detected.

## 4. Discussion

In this paper, we present our I2CU system architecture for real-time monitoring and visual assessment of patients admitted into the ICU. The primary components of the I2CU system include a cart-mounted with multiple sensors that record data which include cameras, wearables, and environmental sensors. The cart also include software modules to control the data collection, run ML inference, and messaging queue. The second component is the server, which receives the data from multiple carts collecting data from ICU rooms and hosts a web application to control the carts remotely. The pervasive cart system consists of an intelligent ML engine with trained ML models performing real-time inference on data collected from multiple sensors. The data from the sensors and the machine learning inference are transferred to the central server using a message queuing tool. Post receiving the complete data from a patient, the data is further curated, cleaned, annotated, and used for training models. ML models are trained on the central server using NVIDIA GPUs, then quantized and deployed into the real-time monitoring system.

The key strength of our I2CU system is the availability of multiple modalities of data. Various forms of data constituting EHR data which include patient demographics, encounters, medication, laboratory tests, and data collected from the wearables, environmental, and vision sensors, provide information that serves as digital biomarkers for predictive modeling. When combined, they can provide a comprehensive picture of the patient's condition by complementing each other. Shickel et al. [46] augmented EHR data with wrist-worn accelerometer data to predict successful or unsuccessful discharge from the ICU and showed marginal improvement in model performance with augmentation. Davoudi et al. [47], in the Intelligent ICU pilot study, showed a statistically significant difference between delirious and non-delirious in extremity movements, head pose, noise and light exposure, and visitation using multiple modalities of data.

Although our I2CU system shows great promise for real-time ICU monitoring of patient clinical status, it has limitations. The cameras to record ICU need to be placed on top of the cart. Therefore, there is a limitation on the weight of the cameras for the stability of the cart. Moreover, the RGB camera needs to be zoomed onto the patient's face, which requires cameras with pan, tilt, and zoom (PTZ) functionalities. With weight limitations and required PTZ functionality, we had to compromise on the camera resolution, which impacted the quality of RGB data in many videos. We use depth cameras for mobility assessments; by the inherent nature of depth images, it is not possible to know who the patient is if multiple people are present in an image. Also, depth cameras, in many cases, cannot resolve a patient lying in bed.

In the current state of our I2CU system, we have developed dedicated deep-learning models for each modality of the sensor data. In the future, we plan to build a model capable of taking input data from different modalities and making an inference overcoming the limitations of each modality by complementing each other. For example, depth images cannot resolve a patient lying in bed and identify a patient among multiple people, while accelerometry data does not have annotated label. A combination of depth and accelerometry can provide a holistic view of patient mobility. We will make our web-based annotation tools available for non-commercial use. We will develop the active learning to be able to run in real-time, which outputs a score on whether collected data should be annotated. A future version of our system can perform real-time vision analysis and store only the video stream, which requires annotation. This approach will also reduce the amount of stored data per patient.

## Acknowledgement

A.B, P.R., and T.B. were supported by NIH/NINDS R01 NS120924, NIH/NIBIB R01 EB029699. P.R. was also supported by NSF CAREER 1750192.